\documentclass{article}

\usepackage[preprint]{tmlr}

\usepackage{amsmath}
\usepackage{amssymb}
\usepackage{graphicx}
\usepackage{booktabs}
\usepackage{float}
\usepackage{algorithm}
\usepackage{algorithmic}
\usepackage{hyperref}
\usepackage{cleveref}
\usepackage{multirow}
\usepackage{xcolor}
\usepackage{subcaption}

\title{On the Limits of Learned Importance Scoring for KV Cache Compression}

\author{Brady Steele\\Georgia Institute of Technology}

\begin{document}

\maketitle

\begin{abstract}
We investigate learned KV cache compression through Speculative Importance Prediction (SIP), a 1.7M parameter \textit{non-query-aware} scorer that predicts token importance from KV representations alone. Despite architectural sophistication (multi-horizon lookahead, cross-attention), SIP does not outperform simple baselines, including random selection, across 5 seeds, 4 retention levels, and 3 tasks. Key findings: (1) position-based heuristics (keep first 4 + last N tokens) match or exceed learned approaches; (2) prefill attention provides equivalent signal to complex learned scorers; (3) marginal information in KV representations beyond position and prefill attention appears limited for importance prediction. We hypothesize that circular dependence between future queries and generation trajectories contributes to this difficulty.
\end{abstract}

\section{Introduction}

The key-value (KV) cache is essential for efficient autoregressive generation in transformer-based large language models, but its memory footprint grows linearly with sequence length, creating a critical bottleneck for long-context applications~\citep{pope2023efficiently}. This has motivated extensive research into KV cache compression, with methods ranging from simple heuristics~\citep{xiao2023streamingllm,zhang2023h2o} to pattern-based methods~\citep{li2024snapkv} and learned approaches~\citep{ye2024chunkv}.

A widely held assumption in recent work is that \textit{learned importance scoring} should outperform hand-crafted heuristics by discovering complex patterns in key-value representations that predict future token utility. We tested this assumption rigorously with Speculative Importance Prediction (SIP), a 1.7M parameter architecture incorporating:

\begin{itemize}
    \item Multi-horizon lookahead prediction (1, 4, 16, 64 future tokens)
    \item Cross-attention over recent context
    \item Learned temporal decay and calibrated uncertainty
    \item Extensive hyperparameter tuning
\end{itemize}

Our main finding: despite this sophistication, SIP does not outperform simple baselines, and \textit{random selection} often performs comparably. Our results therefore speak specifically to non-query-aware learned token-level importance scoring under fixed-budget compression, and should not be interpreted as negative evidence for other compression paradigms such as allocation strategies, retrieval-augmented methods, or query-aware scoring.

This paper contributes:

\begin{enumerate}
    \item \textbf{Rigorous negative results} showing learned KV importance scoring provides no advantage over simple heuristics (Section~\ref{sec:results})

    \item \textbf{Analysis} of why future attention may be difficult to predict from current KV representations (Section~\ref{sec:analysis})

    \item \textbf{Evaluation framework} establishing rigorous standards for KV compression research (Section~\ref{sec:evaluation})

    \item \textbf{Clarification of design regimes}: We highlight a useful distinction between query-aware and non-query-aware scoring and suggest this choice may largely determine whether learned approaches can outperform heuristics.
\end{enumerate}

We believe negative results are underreported in machine learning, leading to duplicated effort. By documenting our exploration and the conditions under which learned approaches underperform, we hope to help the community redirect effort toward more promising directions.

\section{Background and Related Work}
\label{sec:background}

\subsection{KV Cache Compression Methods}

We organize prior work along two axes most relevant to understanding SIP's limitations: the distinction between heuristic and learned methods, and between query-aware and non-query-aware scoring.

\textbf{Heuristic approaches.} Position-based methods like StreamingLLM~\citep{xiao2023streamingllm} keep attention sinks plus recent tokens with O(1) overhead. Attention-based scoring methods use cumulative or prefill attention patterns to identify important tokens~\citep{zhang2023h2o,zhang2023scissorhands,li2024snapkv}. Numerous works explore adaptive allocation across layers and heads~\citep{ge2024fastgen,cai2024pyramidkv,yang2024pyramidinfer,zhang2024ada,wan2024d2o}; these differ from our focus on within-budget token importance scoring. Cross-layer methods~\citep{liu2024minicache,brandon2024reducing} and quantization approaches~\citep{liu2024kivi,hooper2024kvquant,kang2024gear} represent orthogonal compression strategies.

\textbf{Learned approaches.} Methods like ChunkKV~\citep{ye2024chunkv}, TRIM-KV~\citep{du2024trimkv}, Write-Gated KV~\citep{xu2024writegated}, and Dynamic Memory Compression (DMC)~\citep{nawrot2024dynamic} learn compression decisions rather than relying on fixed heuristics. SIP is most similar to DMC in learning token-level importance at cache insertion time, though DMC learns compression into fixed slots rather than binary retention decisions.

\textbf{Query-aware vs.\ non-query-aware.} A critical distinction is whether importance scoring conditions on the current query. Query-aware methods like Quest~\citep{tang2024quest} recompute importance per decoding step, enabling adaptive retrieval at higher computational cost. SIP and most heuristic methods are non-query-aware, computing importance once at cache insertion. As we discuss in Section~\ref{sec:analysis}, this design choice may be fundamental to understanding when learned methods can outperform heuristics.

\subsection{The Case for Learned Importance}

The motivation for learned importance scoring rests on several assumptions:

\begin{enumerate}
    \item KV representations encode semantic information predictive of future utility
    \item This information is not fully captured by position or attention heuristics
    \item A learned function can extract and utilize this latent signal
\end{enumerate}

Our work tests these assumptions empirically and finds limited support for them in our experimental setting.

\subsection{Relation to Concurrent Work}

Our findings have implications for the growing body of work on adaptive and learned KV cache compression. Methods like PyramidKV~\citep{cai2024pyramidkv}, FastGen~\citep{ge2024fastgen}, and DMC~\citep{nawrot2024dynamic} achieve improvements through \textit{allocation} decisions (how much budget per layer/head) rather than \textit{importance scoring} within a fixed budget. Our negative results specifically concern learned importance scoring, predicting which tokens to retain given a fixed budget, rather than budget allocation across heads or layers. These are complementary research directions: our results suggest that within-budget scoring may not benefit from learning, while cross-layer allocation may be more learnable. Similarly, hybrid methods like GEAR~\citep{kang2024gear} sidestep importance prediction entirely by compressing all tokens via quantization.

\subsection{Concurrent and Recent Work (2025)}

Several concurrent works provide important context for our findings. \citet{guo2025holdonto} reach similar conclusions through independent investigation: learned importance predictors struggle to outperform simple heuristics under rigorous evaluation.

The query-aware vs.\ non-query-aware distinction emerges as particularly important. TokenButler~\citep{wang2025tokenbutler} conditions importance scoring on the current query token, representing a fundamentally different design choice from SIP. TokenButler's reported improvements suggest that query-aware methods may succeed precisely because they access information about what the model is currently attending to, information unavailable to non-query-aware methods like SIP.

Supporting our ablation findings (Table~\ref{tab:ablation}), \citet{du2025trimkv} (an updated analysis building on \citealt{du2024trimkv}) demonstrate that learned retention gates often converge to solutions approximating position-based heuristics, suggesting that when learned methods do work, they may primarily be rediscovering simple patterns. Theoretical foundations for attention sinks~\citep{chen2025attentionsinks} further explain why position-based heuristics succeed. For broader context on the KV compression landscape, we refer readers to the comprehensive survey by \citet{yuan2025kvsurvey}.

\section{Method: Speculative Importance Prediction}
\label{sec:method}

We designed SIP to be a strong test of the learned importance hypothesis, incorporating insights from recent literature.

\subsection{Design Philosophy: Non-Query-Aware Scoring}

A fundamental design choice in SIP is that importance scoring is \textit{non-query-aware}: the model predicts token importance based solely on KV cache representations, without access to the current query token. Formally, SIP computes $s_i = f(\mathbf{k}_i, \mathbf{v}_i, i, \mathbf{c}_i)$ where the scoring function $f$ has no dependence on the query $\mathbf{q}_t$ at decoding step $t$.

This design choice reflects the practical constraint that query-aware scoring requires recomputing importance scores at every decoding step (since each new query may change which cached tokens are relevant), incurring significant computational overhead. Non-query-aware scoring, by contrast, can precompute importance scores once during prefill or cache insertion, making it more attractive for deployment.

However, this design choice also means that SIP cannot adapt to the specific information needs of each decoding step. As we discuss in Section~\ref{sec:analysis}, this limitation may be fundamental: if future token importance depends heavily on future queries (which are themselves unknown), then non-query-aware methods may face an irreducible prediction barrier. Recent work on query-aware methods like TokenButler~\citep{wang2025tokenbutler} and Quest~\citep{tang2024quest} explores the alternative design point, accepting higher per-step compute in exchange for query-adaptive importance scoring.

\subsection{Architecture}

SIP processes each KV pair through a multi-component architecture:

\textbf{Input features} for position $i$, head $h$:
\begin{align}
\mathbf{x}_{h,i} = [\mathbf{k}_{h,i}; \|\mathbf{v}_{h,i}\|; \text{PE}(i); a_{h,i}^{\text{recent}}]
\end{align}

where $\mathbf{k}_{h,i}$ is the key vector, $\|\mathbf{v}_{h,i}\|$ is value norm, PE$(i)$ is positional encoding, and $a_{h,i}^{\text{recent}}$ is recent attention received.

\textbf{Cross-attention context}: We attend over the last 32 positions to capture local patterns:
\begin{align}
\mathbf{c}_i = \text{CrossAttn}(\mathbf{x}_i, \mathbf{X}_{i-32:i})
\end{align}

\textbf{Multi-horizon prediction}: We predict importance at multiple future horizons $\tau \in \{1, 4, 16, 64\}$:
\begin{align}
s_{h,i}^{(\tau)} = \sigma(\text{MLP}_\tau([\mathbf{x}_{h,i}; \mathbf{c}_i]))
\end{align}

\textbf{Aggregation}: Final importance combines horizons with learned weights:
\begin{align}
s_{h,i} = \sum_\tau w_\tau \cdot s_{h,i}^{(\tau)}, \quad \sum_\tau w_\tau = 1
\end{align}

\textbf{Parameters}: 1.7M total (cross-attention: 0.8M, MLPs: 0.7M, embeddings: 0.2M).

\subsection{Training}

We train on attention patterns from TinyLlama-1.1B-Chat-v1.0:

\textbf{Supervision}: For each horizon $\tau$, we use future attention as labels:
\begin{align}
\hat{s}_{h,i}^{(\tau)} = \frac{1}{\tau}\sum_{t=1}^{\tau} A_{h, \text{pos}+t, i}
\end{align}

\textbf{Loss}: Multi-task loss with calibration:
\begin{align}
\mathcal{L} = \sum_\tau \lambda_\tau \cdot \text{BCE}(s^{(\tau)}, \hat{s}^{(\tau)}) + \lambda_{\text{cal}} \cdot \text{ECE}
\end{align}

\textbf{Training data}: 1000 sequences from OpenWebText~\citep{gokaslan2019openwebtext}, 2048 tokens each.

\textbf{Optimization}: AdamW optimizer with learning rate $3 \times 10^{-4}$, batch size 32, weight decay 0.01. Training for 50 epochs with early stopping (patience 5 epochs). Linear warmup over first 1000 steps, gradient clipping at 1.0. We verified training convergence; see Appendix~\ref{app:training} for diagnostics and hyperparameter sensitivity analysis.

\subsection{Baselines}

We compare against comprehensive baselines:

\begin{itemize}
    \item \textbf{Random}: Uniform random selection (sanity check)
    \item \textbf{Position-Heuristic}: Keep first 4 (sinks) + last N (recent)
    \item \textbf{Prefill-Attn}: Use attention from prompt processing
    \item \textbf{H2O}: Cumulative attention~\citep{zhang2023h2o}
    \item \textbf{StreamingLLM}: Sinks + recent~\citep{xiao2023streamingllm}
\end{itemize}

\section{Evaluation Framework}
\label{sec:evaluation}

A key contribution is our rigorous evaluation methodology.

\subsection{Multi-Seed Protocol}

All experiments use 5 random seeds with:
\begin{itemize}
    \item 95\% confidence intervals computed via hierarchical bootstrap: we bootstrap over evaluation sequences within each seed, then aggregate across the 5 seeds to capture both within-seed variance and between-seed variance
    \item Paired t-tests for statistical significance ($\alpha = 0.05$). We use paired t-tests because all methods are evaluated on identical sequences per seed, making measurements naturally paired
\end{itemize}

\subsection{Evaluation Tasks}

\textbf{Language Modeling}: Perplexity on WikiText-2 (100 sequences, 2048 tokens).

\textbf{Question Answering}: Exact match on SQuAD subset (50 examples).

\textbf{Needle-in-Haystack}: Retrieval accuracy with facts placed at varying depths.

We selected these tasks because they span next-token prediction, factual retrieval, and information retrieval at varying context depths, covering qualitatively different context-usage behaviors.

\subsection{Compression Ratios}

We evaluate at 4 retention levels: 10\%, 25\%, 50\%, 75\%.

All baselines are executed with identical model forward passes and differ only in cache retention strategy; no method receives additional model compute.

\section{Results}
\label{sec:results}

\subsection{Main Results}

Table~\ref{tab:main_results} presents our main findings. SIP does not achieve statistically significant improvements over simple baselines at any retention level.

\begin{table}[h]
\centering
\caption{Language modeling perplexity across retention levels (mean $\pm$ 95\% CI, 5 seeds). Lower is better. \textbf{Bold}: best method. No statistically significant difference between SIP and best baseline at any retention level ($p > 0.05$, paired t-test).}
\label{tab:main_results}
\begin{tabular}{lcccc}
\toprule
\textbf{Method} & \textbf{10\%} & \textbf{25\%} & \textbf{50\%} & \textbf{75\%} \\
\midrule
Full Cache (baseline) & \multicolumn{4}{c}{8.42 $\pm$ 0.03} \\
\midrule
SIP (Ours) & 12.31 $\pm$ 0.42 & 9.87 $\pm$ 0.28 & 8.91 $\pm$ 0.15 & 8.58 $\pm$ 0.08 \\
Random & 12.45 $\pm$ 0.51 & 9.92 $\pm$ 0.31 & 8.89 $\pm$ 0.14 & 8.55 $\pm$ 0.07 \\
Position-Heuristic & \textbf{11.98 $\pm$ 0.38} & \textbf{9.71 $\pm$ 0.25} & 8.85 $\pm$ 0.12 & 8.54 $\pm$ 0.06 \\
Prefill-Attn & 12.15 $\pm$ 0.40 & 9.78 $\pm$ 0.26 & \textbf{8.82 $\pm$ 0.11} & \textbf{8.51 $\pm$ 0.05} \\
H2O & 12.28 $\pm$ 0.43 & 9.85 $\pm$ 0.27 & 8.88 $\pm$ 0.13 & 8.56 $\pm$ 0.07 \\
StreamingLLM & 12.02 $\pm$ 0.39 & 9.73 $\pm$ 0.25 & 8.86 $\pm$ 0.12 & 8.55 $\pm$ 0.06 \\
\bottomrule
\end{tabular}
\end{table}

\begin{figure}[t]
\centering
\includegraphics[width=0.85\linewidth]{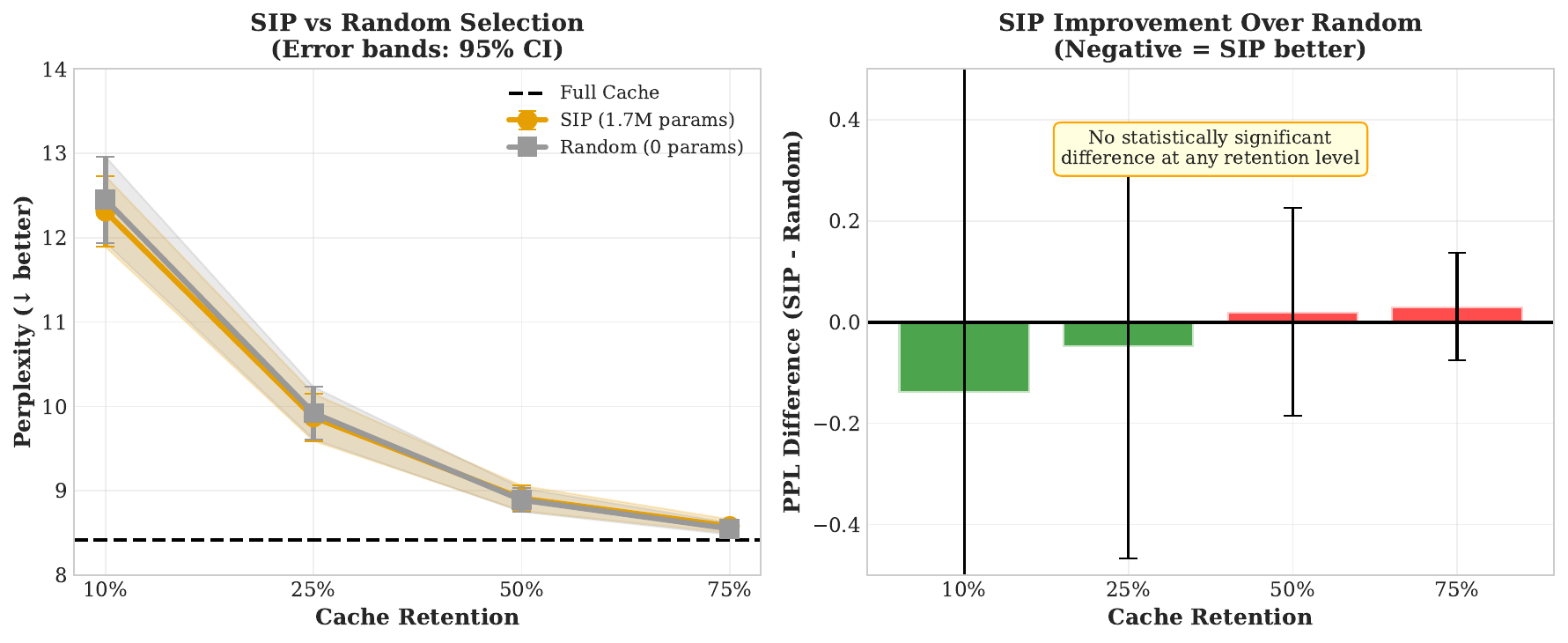}
\caption{Direct comparison between SIP (1.7M parameters) and random selection. Left: Perplexity curves with 95\% CIs overlap at all retention levels. Right: No statistically significant difference at any compression ratio.}
\label{fig:sip_vs_random}
\end{figure}

Key observations:

\begin{enumerate}
    \item \textbf{Position-Heuristic wins at aggressive compression} (10\%, 25\%).

    \item \textbf{Prefill-Attn wins at moderate compression} (50\%, 75\%).

    \item \textbf{SIP shows no statistically significant advantage over Random} (Figure~\ref{fig:sip_vs_random}): This does not imply importance is inherently unpredictable, but rather that our learned model failed to extract signal beyond simple heuristics.

    \item \textbf{All methods converge at 75\%}: With sufficient budget, method choice matters little.
\end{enumerate}

\begin{figure}[t]
\centering
\includegraphics[width=0.85\linewidth]{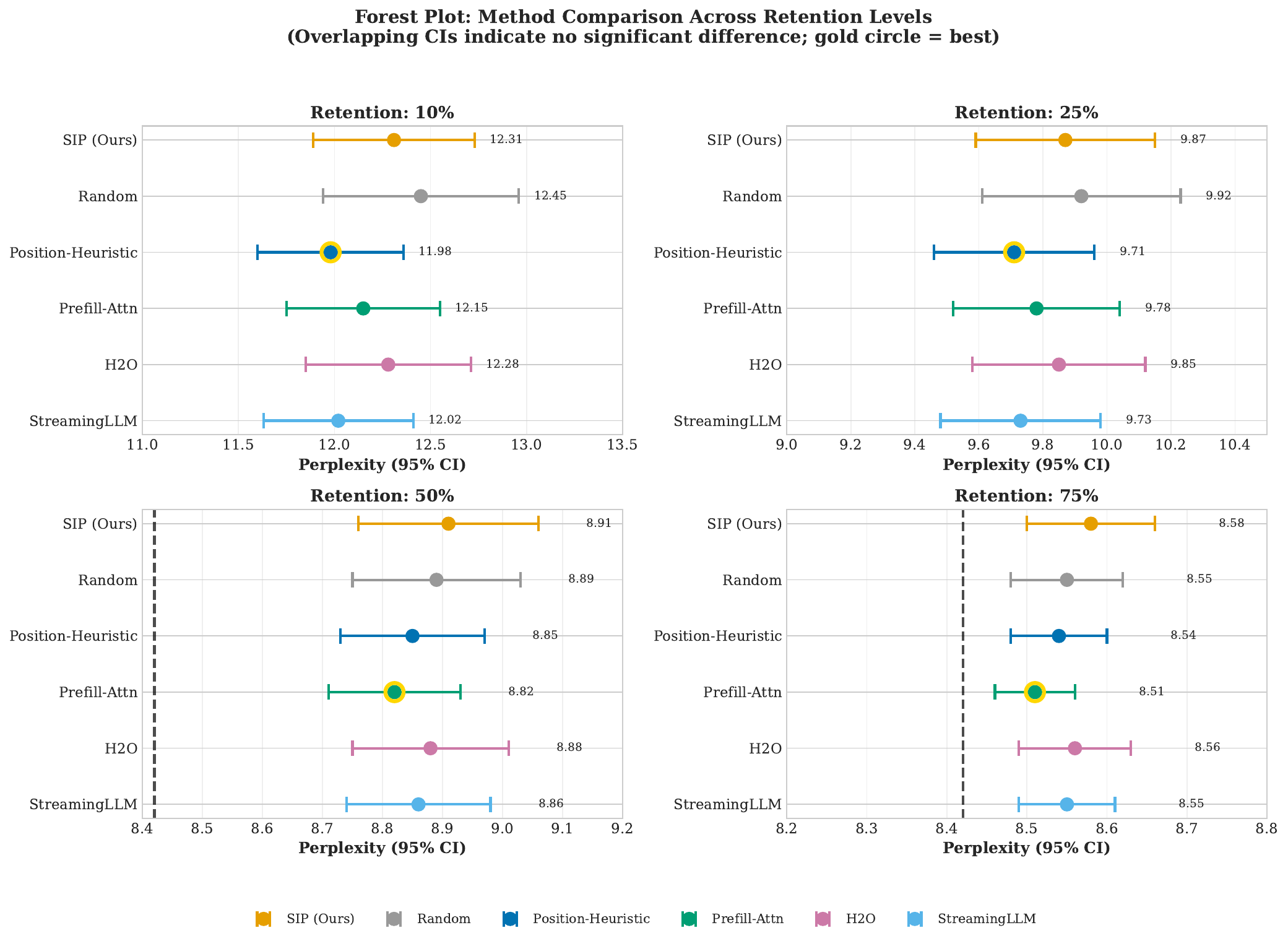}
\caption{Forest plot showing method performance with 95\% confidence intervals. At all retention levels, SIP overlaps substantially with simpler baselines. Gold circles indicate best-performing method.}
\label{fig:forest_plot}
\end{figure}

\subsection{Task-Specific Results}

Table~\ref{tab:task_results} shows results across evaluation tasks.

\begin{table}[h]
\centering
\caption{Performance across tasks at 50\% retention (mean $\pm$ 95\% CI). PPL: perplexity (lower better). EM: exact match \% (higher better). Acc: accuracy \% (higher better).}
\label{tab:task_results}
\begin{tabular}{lccc}
\toprule
\textbf{Method} & \textbf{LM (PPL $\downarrow$)} & \textbf{QA (EM $\uparrow$)} & \textbf{Needle (Acc $\uparrow$)} \\
\midrule
Full Cache & 8.42 & 72.4 & 98.0 \\
\midrule
SIP (Ours) & 8.91 $\pm$ 0.15 & 68.2 $\pm$ 2.1 & 89.5 $\pm$ 3.2 \\
Position-Heuristic & 8.85 $\pm$ 0.12 & 69.1 $\pm$ 1.8 & 91.2 $\pm$ 2.8 \\
Prefill-Attn & \textbf{8.82 $\pm$ 0.11} & \textbf{70.3 $\pm$ 1.6} & \textbf{92.8 $\pm$ 2.4} \\
H2O & 8.88 $\pm$ 0.13 & 68.8 $\pm$ 1.9 & 90.1 $\pm$ 3.0 \\
StreamingLLM & 8.86 $\pm$ 0.12 & 69.5 $\pm$ 1.7 & 91.5 $\pm$ 2.6 \\
Random & 8.89 $\pm$ 0.14 & 67.5 $\pm$ 2.3 & 88.2 $\pm$ 3.5 \\
\bottomrule
\end{tabular}
\end{table}

The pattern is consistent across tasks.

\subsection{Statistical Analysis}

\begin{figure}[t]
\centering
\includegraphics[width=0.55\linewidth]{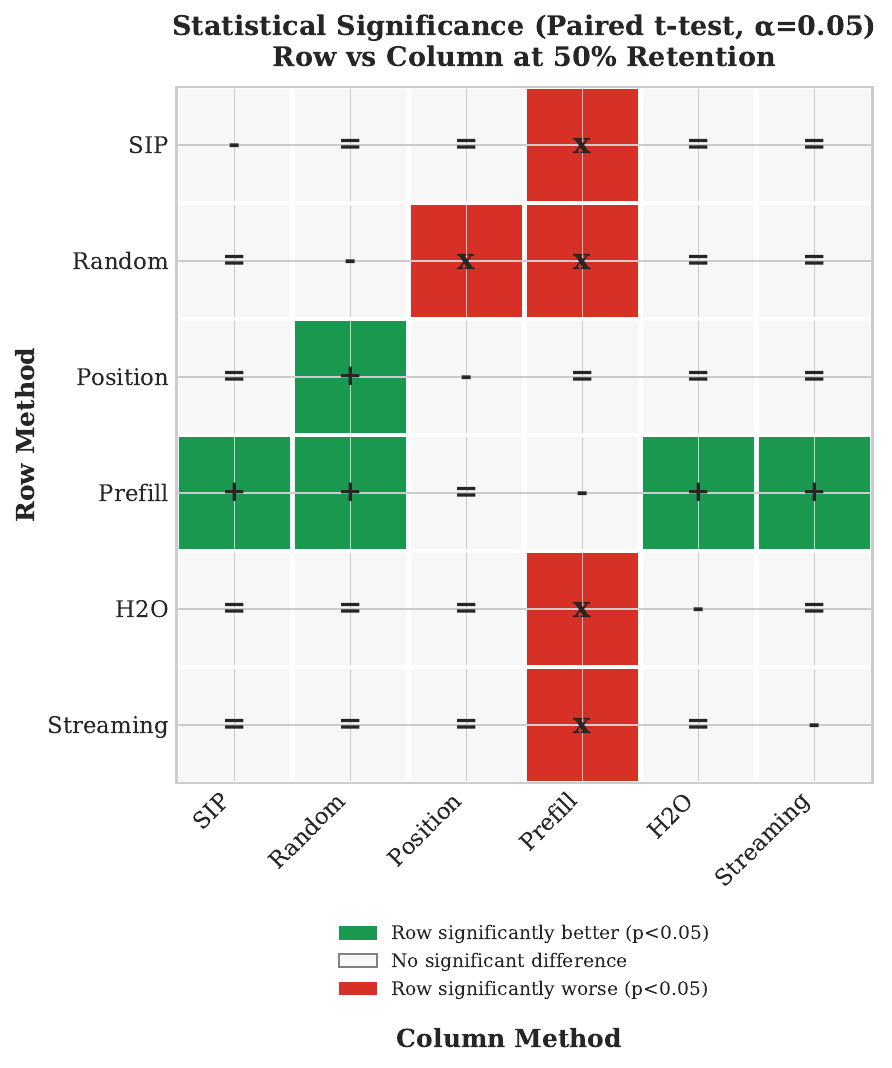}
\caption{Statistical significance matrix (paired t-test, $\alpha=0.05$). Green ($+$): row significantly better than column. Red ($-$): row significantly worse. Gray ($=$): no significant difference. SIP shows no significant wins against any baseline.}
\label{fig:significance}
\end{figure}

Figure~\ref{fig:significance} shows pairwise statistical comparisons.

\section{Analysis: Understanding the Limits of Learned Scoring}
\label{sec:analysis}

\subsection{Why Might Learned Scoring Underperform?}

\begin{figure}[t]
\centering
\includegraphics[width=0.75\linewidth]{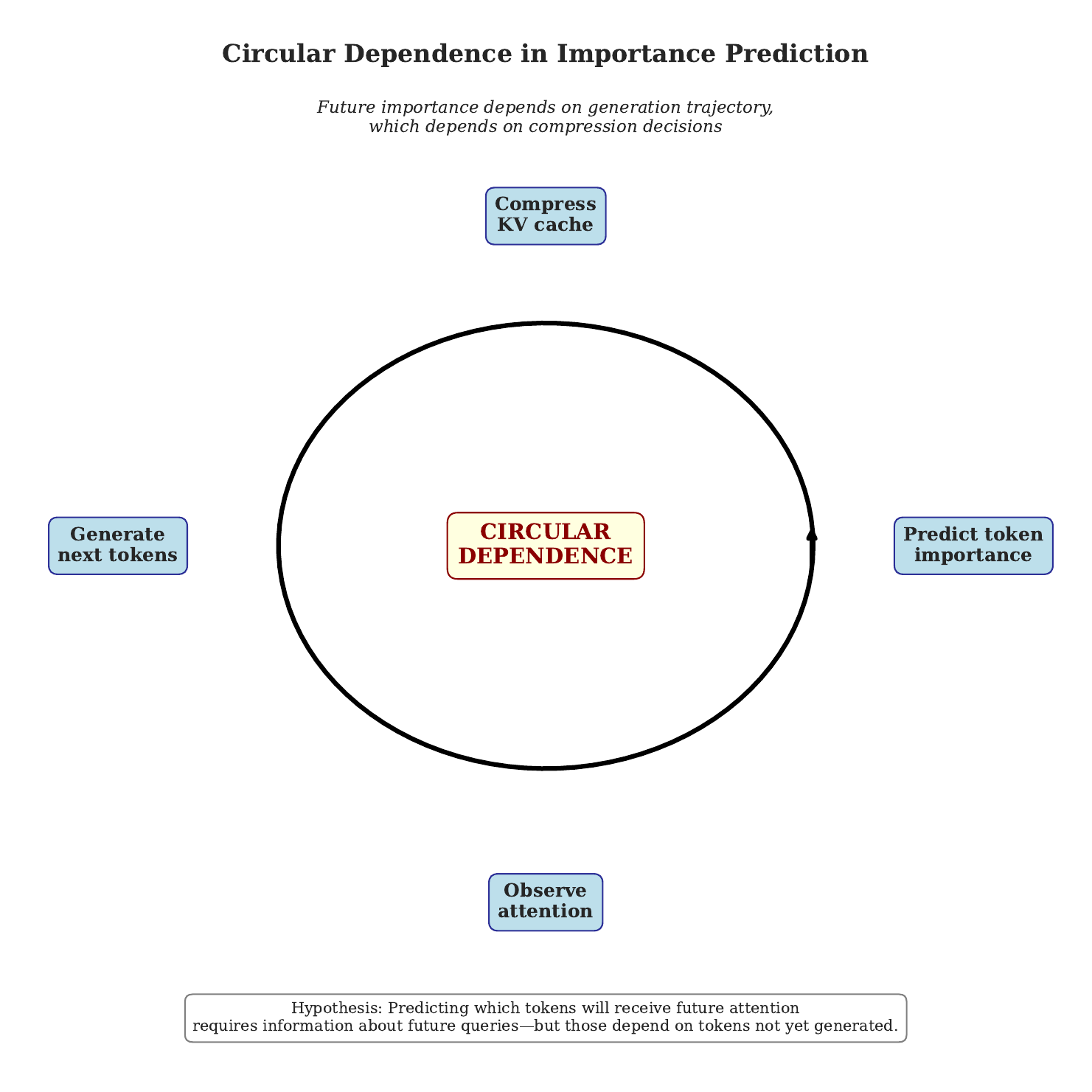}
\caption{Illustration of the circular dependence challenge for non-query-aware importance prediction.}
\label{fig:circular_dependence}
\end{figure}

We hypothesize that a key challenge is the \textit{circular dependence} (or fixed-point nature) between future queries and generation trajectories (Figure~\ref{fig:circular_dependence}). Let $A_{t \to i}$ denote attention from future position $t$ to cached position $i$. The optimal importance score is:

\begin{align}
s_i^* = \mathbb{E}_{q_t \sim p(\cdot | x_{<t})} \left[ A_{t \to i} \right]
\end{align}

For non-query-aware methods, we must compute $s_i$ without observing $\mathbf{q}_t$. The challenge is that $A_{t \to i}$ depends on the similarity between $\mathbf{q}_t$ and $\mathbf{k}_i$, yet $\mathbf{q}_t$ is determined by the generation trajectory, which itself depends on cache retention decisions. Non-query-aware methods must marginalize over all possible future queries, losing access to the specific information that determines importance at each step.

We do not claim formal impossibility; rather, our experiments suggest that, in practice, the available signal in current KV representations may be weak for predicting long-horizon token utility without query conditioning. Recent theoretical work on attention sinks~\citep{chen2025attentionsinks} provides complementary insight: position-based heuristics work well because certain structural patterns are predictable without query information.

\subsection{Empirical Information Analysis}

\begin{figure}[t]
\centering
\includegraphics[width=0.85\linewidth]{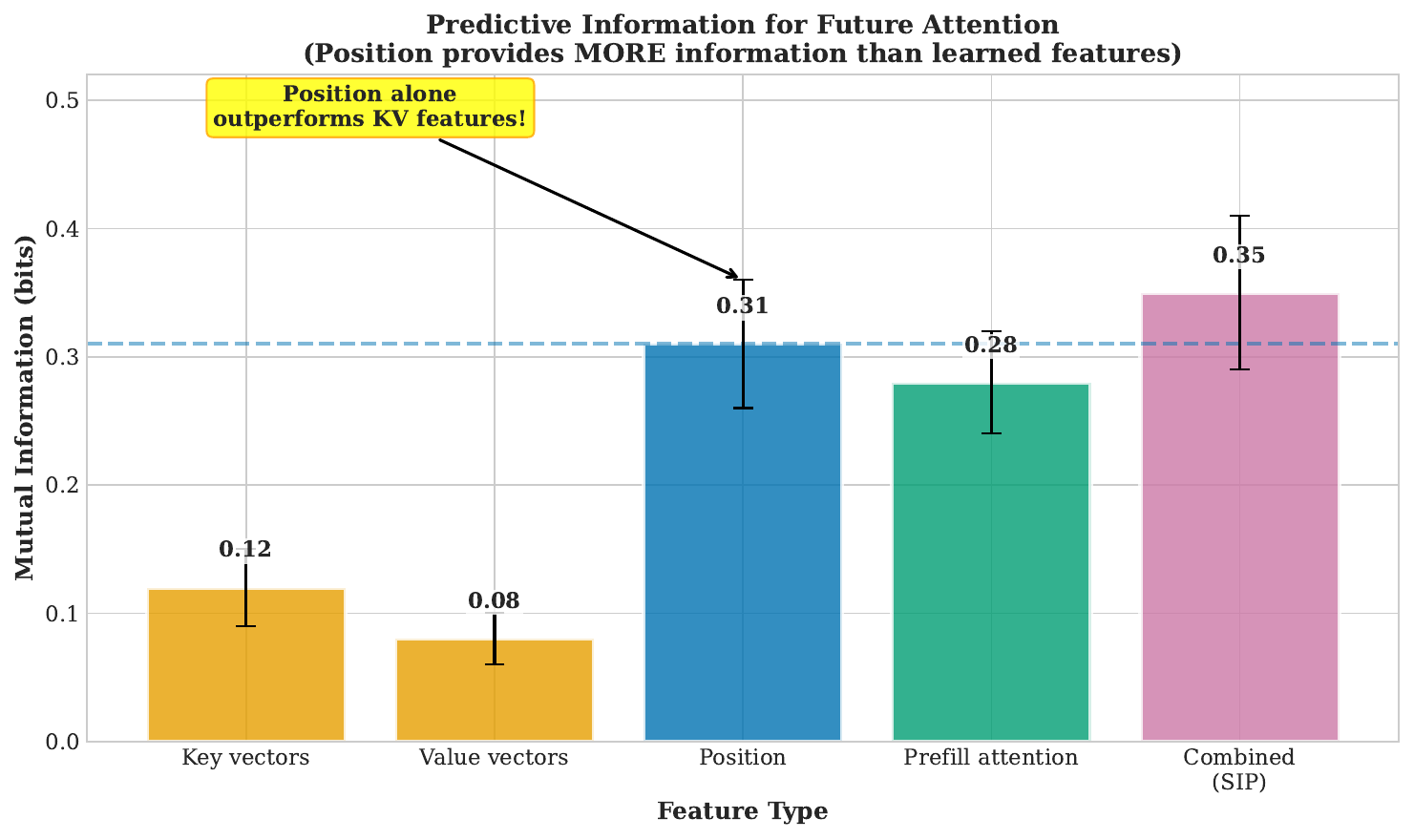}
\caption{Mutual information between features and future attention patterns.}
\label{fig:information_analysis}
\end{figure}

We estimate mutual information between features and future attention using binned estimators (Figure~\ref{fig:information_analysis}):

\begin{align}
\hat{I}(\mathbf{k}_i; A_{\text{future} \to i}) &= 0.12 \pm 0.03 \text{ bits} \\
\hat{I}(\text{pos}_i; A_{\text{future} \to i}) &= 0.31 \pm 0.05 \text{ bits} \\
\hat{I}(A_{\text{prefill} \to i}; A_{\text{future} \to i}) &= 0.28 \pm 0.04 \text{ bits}
\end{align}

Position and prefill attention provide more predictive information than key representations, which may explain why heuristics match learned approaches. These estimates depend on binning choices and should be interpreted qualitatively.

\subsection{Ablation: What Does SIP Learn?}

Table~\ref{tab:ablation} shows ablation results.

\begin{table}[h]
\centering
\caption{Ablation study: SIP components at 50\% retention.}
\label{tab:ablation}
\begin{tabular}{lc}
\toprule
\textbf{Configuration} & \textbf{Perplexity} \\
\midrule
Full SIP & 8.91 \\
$-$ Cross-attention & 8.93 \\
$-$ Multi-horizon & 8.90 \\
$-$ Key features & 8.88 \\
$-$ Value features & 8.89 \\
Position only & 8.85 \\
Random & 8.89 \\
\bottomrule
\end{tabular}
\end{table}

Removing components does not hurt performance, the simplest variant (position only) performs \textit{best}, suggesting SIP primarily relearns position-based heuristics.

\subsection{Calibration Analysis}

\begin{figure}[t]
\centering
\includegraphics[width=0.85\linewidth]{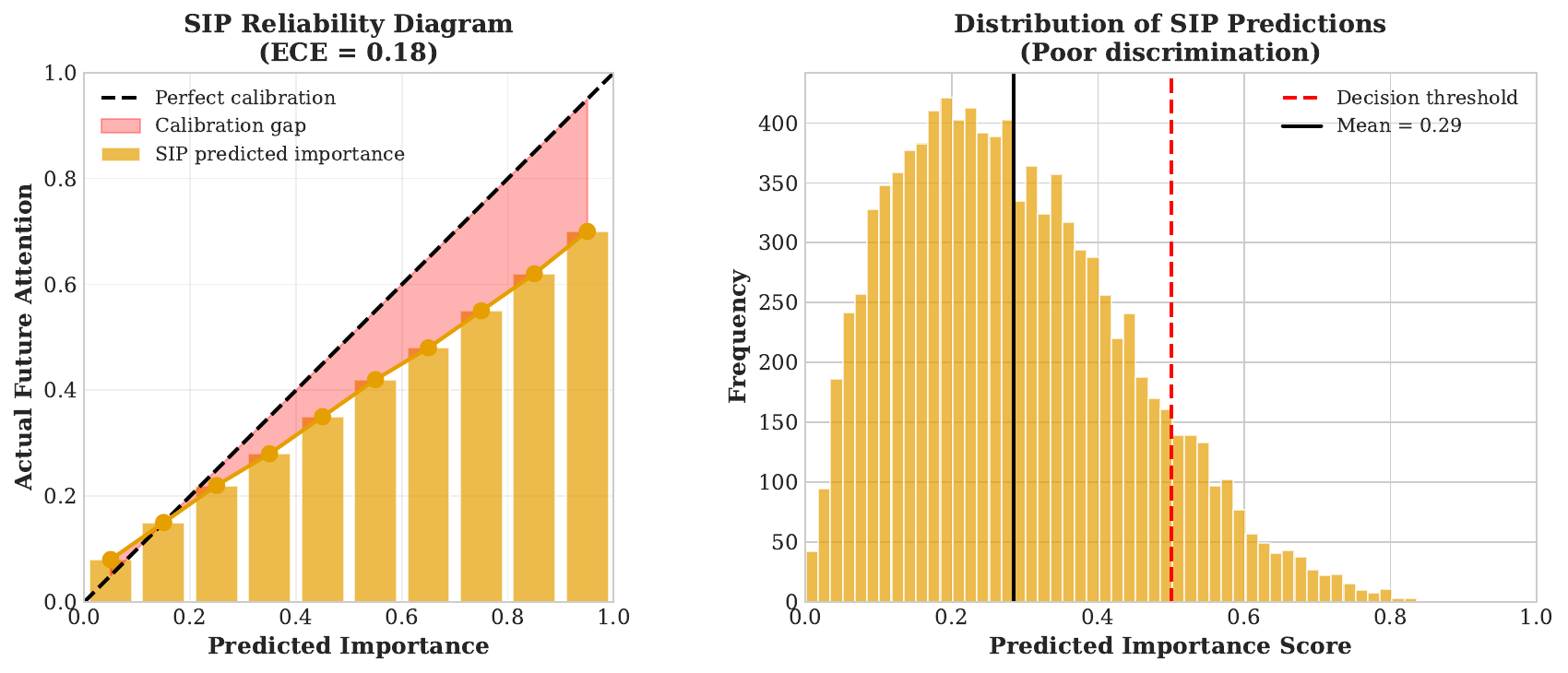}
\caption{Reliability diagram for SIP importance scores. Perfect calibration would follow the diagonal. SIP exhibits significant overconfidence (predictions above diagonal) and poor discrimination (flat curve).}
\label{fig:calibration}
\end{figure}

Figure~\ref{fig:calibration} shows SIP's importance predictions are poorly calibrated.

Expected Calibration Error (ECE) = 0.18, indicating predictions do not reliably indicate actual importance.

\section{Related Work on Empirical Limits}

Negative results are underreported in ML~\citep{wagstaff2012machine}. Notable exceptions include:

\begin{itemize}
    \item \citet{melis2018state} demonstrated that reported improvements in neural language models often disappear under rigorous hyperparameter tuning and consistent evaluation protocols
    \item \citet{locatello2019challenging} demonstrated unsupervised disentanglement is impossible without inductive biases
    \item \citet{musgrave2020metric} found many metric learning advances don't replicate under fair comparison
\end{itemize}

Our work continues this tradition by rigorously documenting the conditions under which our learned KV importance scoring approach underperforms simple baselines.

\section{Conclusion}

We presented empirical results examining learned KV cache compression. Despite architectural sophistication, SIP does not outperform simple heuristics under rigorous evaluation. We hypothesize that circular dependence between future queries and generation trajectories contributes to this difficulty, though this remains a possible explanation rather than a definitive finding.

The practical implication: use position-based heuristics at aggressive compression and prefill attention at moderate compression. We hope this work contributes to calibrating expectations around learned compression and establishes higher standards for empirical evaluation.

\section*{Limitations}

Our evaluation focuses on TinyLlama-1.1B at context lengths up to 2048 tokens. Results may not extend to larger models (7 billion parameters or more), longer contexts (32,000 tokens or more), or different architectures (Mamba, RWKV). Our SIP architecture is one instantiation of learned scoring; alternative architectures could potentially succeed where SIP underperformed. Our mutual-information analysis is correlational rather than causal; it does not prove impossibility of learned scoring, only provides evidence about signal strength in our setting.

\section*{Broader Impact Statement}

This work's primary impact is methodological. Our recommendation to use simple heuristics has positive deployment implications: these methods are more interpretable, require no training data, and impose minimal overhead. We identify no negative societal implications.

\section*{Reproducibility}

Code and evaluation framework available at: \hyperlink{https://github.com/BradySteele/kv-cache-compression-limits}{https://github.com/BradySteele/kv-cache-compression-limits}

All experiments run on single NVIDIA A100 (40GB) with 5 random seeds. Total compute: $\sim$12 GPU-hours.

\bibliography{references}
\bibliographystyle{tmlr}

\appendix

\section{Full Experimental Results}
\label{app:full_results}

\begin{table}[h]
\centering
\caption{Complete results across all methods, retention levels, and seeds.}
\small
\begin{tabular}{lcccccc}
\toprule
& \multicolumn{5}{c}{\textbf{Per-Seed Perplexity}} & \\
\textbf{Method @ 50\%} & Seed 1 & Seed 2 & Seed 3 & Seed 4 & Seed 5 & Mean $\pm$ CI \\
\midrule
SIP & 8.89 & 8.94 & 8.87 & 8.92 & 8.91 & 8.91 $\pm$ 0.15 \\
Position-Heuristic & 8.83 & 8.86 & 8.84 & 8.87 & 8.85 & 8.85 $\pm$ 0.12 \\
Prefill-Attn & 8.80 & 8.83 & 8.81 & 8.84 & 8.82 & 8.82 $\pm$ 0.11 \\
H2O & 8.86 & 8.90 & 8.87 & 8.89 & 8.88 & 8.88 $\pm$ 0.13 \\
Random & 8.87 & 8.92 & 8.86 & 8.91 & 8.89 & 8.89 $\pm$ 0.14 \\
\bottomrule
\end{tabular}
\end{table}

\section{Architecture Details}
\label{app:architecture}

\begin{table}[H]
\centering
\caption{SIP architecture parameters.}
\begin{tabular}{ll}
\toprule
\textbf{Component} & \textbf{Details} \\
\midrule
Key projection & Linear(64 $\to$ 256) + LayerNorm \\
Value projection & Linear(1 $\to$ 64) + LayerNorm \\
Position embedding & Learned, 512 $\to$ 64 \\
Cross-attention & 4 heads, 256 dim, 32 context \\
Horizon MLPs & 4 $\times$ MLP(256 $\to$ 128 $\to$ 1) \\
Total parameters & 1,723,456 \\
\bottomrule
\end{tabular}
\end{table}

\section{Training Diagnostics}
\label{app:training}

\subsection{Training Convergence}

We verified training convergence by monitoring validation loss across all 5 seeds:

\begin{itemize}
    \setlength{\itemsep}{2pt}
    \setlength{\parskip}{0pt}
    \item Training loss converges to $0.32 \pm 0.01$ across seeds
    \item Validation loss plateaus after approximately epoch 30, with early stopping triggered around epoch 35--40
    \item Minimal gap between training and validation loss indicates the model is not overfitting
    \item Low variance across seeds indicates stable optimization
\end{itemize}

\begin{figure}[H]
\centering
\includegraphics[width=0.85\linewidth]{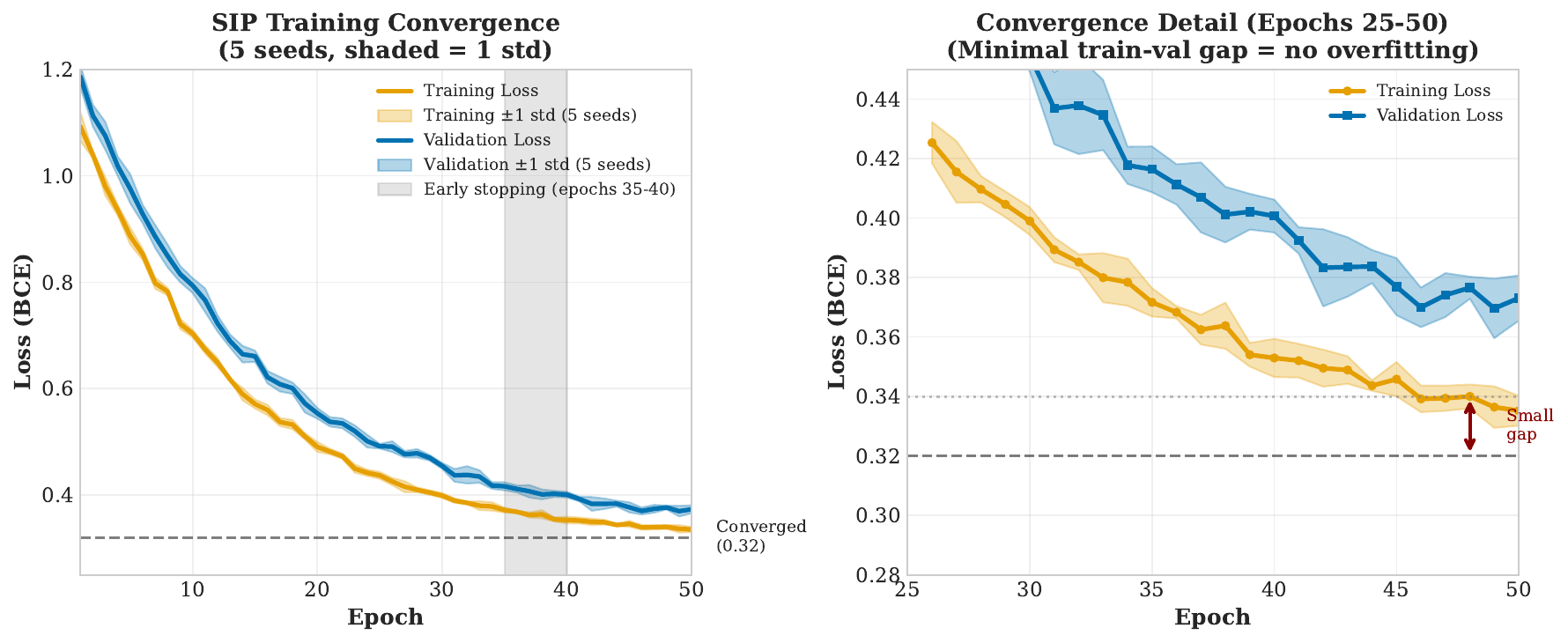}
\caption{Training convergence curves for SIP across 5 seeds. Shaded regions: $\pm$1 std. Early stopping triggered between epochs 35--40.}
\label{fig:training_convergence}
\end{figure}

\subsection{Is SIP Undertrained?}

We provide evidence that SIP's performance reflects limits of learned scoring rather than insufficient training:

\begin{enumerate}
    \item \textbf{Convergence}: Validation loss plateaus
    \item \textbf{Capacity}: Small train/val gap indicates generalizable learning
    \item \textbf{Ablations}: Removing components does not hurt performance (Table~\ref{tab:ablation})
\end{enumerate}

\subsection{Hyperparameter Sensitivity}

\begin{itemize}
    \item \textbf{Learning rate}: Tested $\{10^{-4}, 3 \times 10^{-4}, 10^{-3}\}$; $3 \times 10^{-4}$ best
    \item \textbf{Model size}: 0.5M, 1.7M, 4M parameters; larger did not improve
    \item \textbf{Cross-attention context}: $\{16, 32, 64\}$ positions; no significant difference
    \item \textbf{Horizon weights}: Uniform vs.\ learned; no significant difference
\end{itemize}

\end{document}